\documentclass[runningheads]{llncs}

\usepackage[T1]{fontenc}
\usepackage{graphicx}

\usepackage{xcolor}

\usepackage{amsmath}
\usepackage{xfrac}
\usepackage{gensymb}
\usepackage{fixmath}

\usepackage{array}

\usepackage[hidelinks]{hyperref}

\usepackage{comment}
\usepackage{caption}
\usepackage{subcaption}

\usepackage[english]{babel}
\usepackage[utf8]{inputenc}
\usepackage{vhistory}
\usepackage{enumitem}
\usepackage{longtable}

\usepackage[backend=biber, style=ieee, citestyle=numeric-comp]{biblatex}

\bibliography{bnaic_references}

\title{Bringing the RT-1-X Foundation Model \\ to a SCARA robot}

\date{August 2024}

\author{Jonathan Salzer \and
Arnoud Visser\orcidID{\href{https://orcid.org/0000-0002-7525-7017}{\includegraphics[scale=0.05]{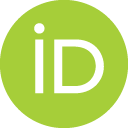}}}}

\institute{
Intelligent Robotics Lab, Universiteit van Amsterdam, NL
\hspace{5em}
\href{https://www.intelligentroboticslab.nl/}{www.intelligentroboticslab.nl}}

\begin{document}

\maketitle

\begin{abstract}
Traditional robotic systems require specific training data for each task, environment, and robot form. While recent advancements in machine learning have enabled models to generalize across new tasks and environments, the challenge of adapting these models to entirely new settings remains largely unexplored. This study addresses this by investigating the generalization capabilities of the RT-1-X robotic foundation model to a type of robot unseen during its training: a SCARA robot from UMI-RTX.

Initial experiments reveal that RT-1-X does not generalize zero-shot to the unseen type of robot. However, fine-tuning of the RT-1-X model by demonstration allows the robot to learn a pickup task which was part of the foundation model (but learned for another type of robot). 
When the robot is presented with an object that is included in the foundation model but not in the fine-tuning dataset, it demonstrates that only the skill, but not the object-specific knowledge, has been transferred.

\keywords{Imitation learning \and Tokenization \and Conditioning.}

\end{abstract}

\section{Introduction}
\label{sec:introduction}

Recent breakthroughs in machine learning and artificial intelligence suggest that training on large, diverse datasets can lead to highly adaptable models, which often exceed the performance of models developed for specific tasks using smaller datasets \cite{open-x-embodiment-collaboration-open-2024}.
Traditional approaches to robotic leaning have required task-specific datasets tailored to each individual task, environment, and robot, limiting the scalability and flexibility of robotic applications. As a result, the field has been exploring more generalizable models that can adapt to new scenarios with minimal retraining. Recent advancements like transformer architectures and robotic foundation models have led to models like Google's RT-1 \cite{brohan_rt-1_2022}, which demonstrate the potential for robots to generalize across various tasks and environments. However, one critical area remains underexplored: the ability of these models to generalize across entirely new robotic embodiments.

\begin{figure}
    \centering
    \includegraphics[width=0.45\linewidth]{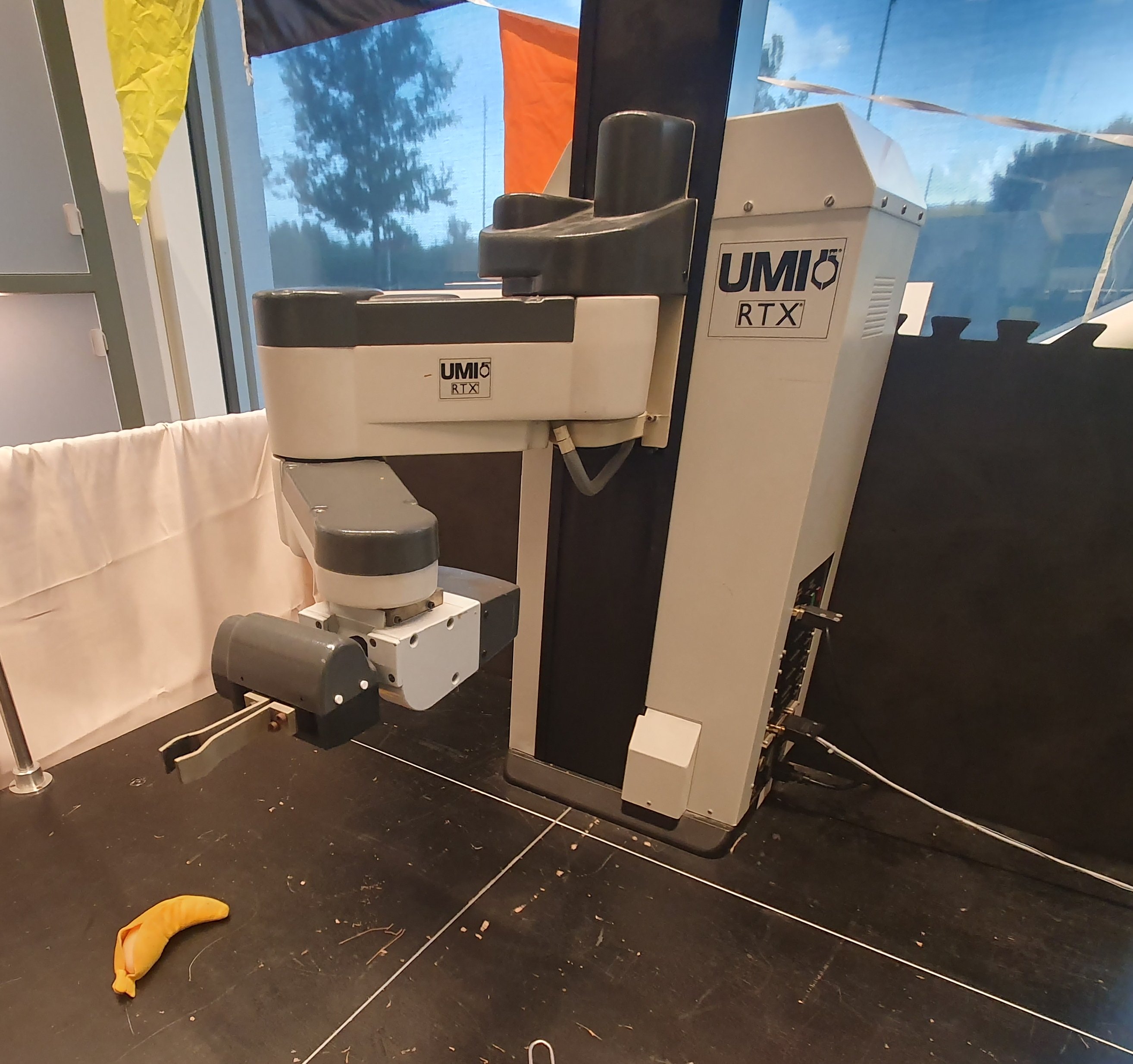}
    \caption{The UMI-RTX robot with an object in its working space.}
    \label{fig:umi-uva}
    \vspace*{-1em}
\end{figure}

Google's RT-1 model is an impressive work, tested on a collection of real-world robotic experiences, where in different institutes a fleet of robots were performing 700 tasks \cite{brohan_rt-1_2022}.  The robots in the training set, such as the Franka, Kuka iiwa, UR5 and the EveryDay robot, can move their end-effector in a spherical working-space. None of the robots in the dataset is of the SCARA  (Selective Compliance Assembly Robot Arm) type. With a SCARA robot the movement of z-axis is decoupled from the movement in the x-y plane, which gives a SCARA robot an kidney shape working-space. As example of such SCARA robot is the classic UMI RTX robot, a robot which is still functional at the University of Amsterdam's Intelligent Robotics Lab, despite its age of nearly 40 years (see Fig.~\ref{fig:umi-uva}).

More details on this particular robot is given in Sec.~\ref{sec:umi-rtx}. The focus of this study is to see if generalization capabilities of the Google's RT-1 model can be extended to an unseen robot, of a complete different type.

\section{Theoretical Background}
\label{sec:background}

This section introduces the three core components that this research is based on: the RT-1 model, the Open X-Embodiment dataset, and the UMI-RTX robot.

The field of machine learning in the robotic domain is  constantly expanding, with new work published very regularly \cite{zheng2024surveyembodiedlearningobjectcentric}. To stay within the scope, this introduction will be organized around RT-1-X and its position within the robotic learning field.

\subsection{RT-1}
The RT-1 model \cite{brohan_rt-1_2022} was presented as a joint effort between Robotics at Google, Everyday Robots, and Google Research, at the end of 2022. 
The purpose of RT-1 is to investigate if it is possible to train a single, capable, multi-task model on data consisting of a wide variety of robotic tasks, and to find out if such a model brings the same benefits observed in other domains, namely zero-shot generalization to new tasks, environments, and objects.

\begin{figure}[ht!]
    \centering
    \includegraphics[width=0.65\linewidth]{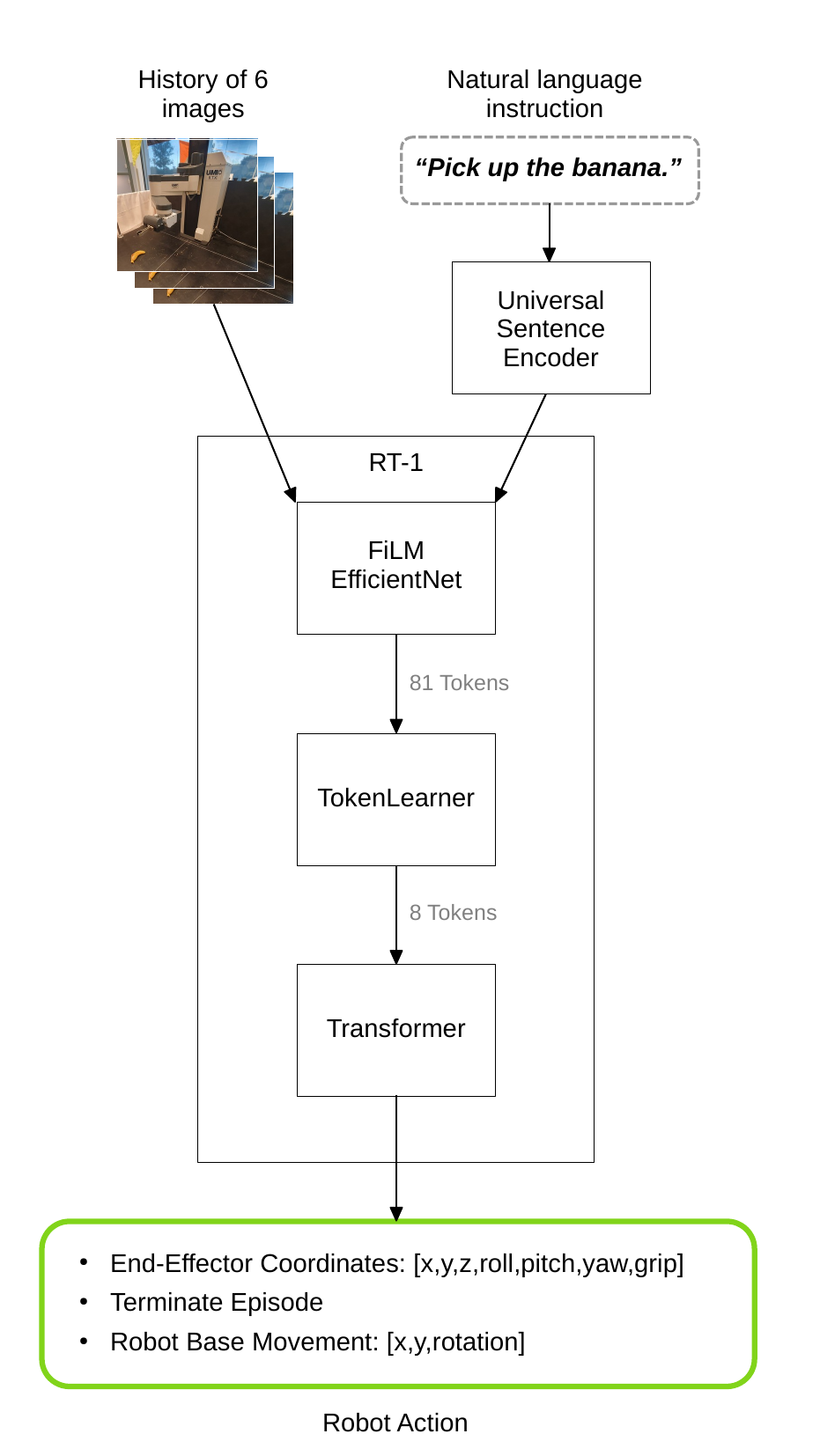}
    \caption{A simplified version of the RT-1 architecture, showing the input of an image history and language instruction, application of the various modules, and the robot action output format.}
    \label{fig:rt1-simplified}
    \vspace*{-1em}
\end{figure}

With RT-1, the authors present a model architecture along with a significant training dataset that fulfills those requirements, as well as demonstrate the success of this model \cite{brohan_rt-1_2022}. The basic input and output of the RT-1 model is illustrated in Fig. \ref{fig:rt1-simplified}: The model takes a natural language instruction, along with a history of six RGB images, as input, and returns a eleven-dimensional action description as output: seven dimensions for arm movement, three dimensions for base movement, and one dimension for terminating an episode. Note that the three dimensions for base movement are irrelevant for this research, as a stationary robot is used, it is only dealt with the seven-dimensional arm movement vector from here onwards.

The following sections describes the several advanced components which are integrated in the (black) box labeled RT-1 in Fig.~\ref{fig:rt1-simplified}.

\subsubsection{Universal Sentence Encoder}
 The natural language instruction input is processed by an Universal Sentence Encoder \cite{cer-etal-2018-universal}. The Universal Sentence Encoder (USE) is a pre-trained model introduced in \citeyear{cer-etal-2018-universal}, designed to convert sentences into high-dimensional embedding vectors. It is specifically targeted towards transfer learning to other NLP tasks, meaning that its goal is to provide a general-purpose embedding that can be used in a wide range of NLP applications. It produces fixed-length embeddings of 512-dimensional vectors, regardless of the input sentence length. The embeddings can be used to determine the semantic similarity between short pieces of text. 

\subsubsection{FiLM}
Feature-wise Linear Manipulation (FiLM) \cite{perez_film_2018} is a conditioning method for neural networks introduced in \citeyear{perez_film_2018}. It works by introducing additional layers to a CNN. FiLM layers carry out simple, feature-wise affine transformations on a network's intermediate features, conditioned on an arbitrary input. This significally alters the CNN's behavior depending on the conditioning input, allowing the overall model to carry out a variety of conditioning tasks. In the case of RT-1, it is used to extract task-relevant image features based on the language instruction. 

\subsubsection{EfficientNet}
The input images are processed by an EfficientNet variant \cite{tan_efficientnet_2019}. EfficientNet is a family of Convolutional Neural Networks (CNNs) introduced in \citeyear{tan_efficientnet_2019}. The key innovation of EfficientNet is a method called compound scaling: 
all three dimensions (depth, width \& resolution) scale uniformly, using a simple yet highly effective compound component. Based on this, the authors develop a family of models (EfficientNet-B0 (5.3M params) to B7 (66M params)), which achieve much better accuracy and efficiency than any previous CNNs. In RT-1, the B3 version of EfficientNet is used. 

\subsubsection{Input Tokenization}
The two input types are now combined. The images are tokenized by passing them through an EfficientNet-B3, pre-trained on ImageNet, which takes six images of resolution 300x300 as input and returns a spatial feature map of shape 9x9x512 per image. This feature map is then flattened into 81 visual tokens, which can be passed to the later layers of the network.

To include the natural language instruction, it is first embedded using the universal sentence encoder, which transforms it to a vector of length 512. This vector is then used as an input to identity-initialized FiLM layers, which are added to the pre-trained EfficientNet to condition the image encoder, to extract task-relevant image features early on. 

The output after the image and instruction tokenization is 81 vision-language tokens per image. To further speed up inference, the number of tokens that the Transformer needs to attend over is further compressed using TokenLearner. From the 81 tokens that come out of the FiLM-conditioned EfficientNet, TokenLearner derives only 8 final tokens.

\subsubsection{TokenLearner}

TokenLearner is a learnable module that takes images as input, determines which parts of the image are "worth processing", and based on that generates a small set of tokens. In RT-1, it works by selecting relevant image tokens based on their information, and passing only important token combinations on to the subsequent Transformer layers \cite{brohan_rt-1_2022}. By doing so, it saves memory and computation by more than half, without impacting classification performance \cite{ryoo_tokenlearner_2021}.  

\subsection{Open X-Embodiment and RT-1-X}
Recent advancements in machine learning have shown that large-scale training on diverse datasets can lead to general-purpose models that even outperform their narrowly targeted counterparts, trained on smaller, task specific data, by leveraging the benefits of positive transfer between domains. Increasingly, the go-to approach to tackle a given narrow task e.g. in vision or NLP, is to adapt a general-purpose model. In the previous section, findings from RT-1 were presented, which show that this approach could also work in robotics. In this domain however it is hard to apply on a larger scale, since as previously discussed, datasets for robotic interaction are hard to come by. Even the largest existing robotic datasets are a fraction of the size of their vision or NLP counterparts, in addition robotic datasets are very often still narrow along some axes of variation, either all collected on a single environment, or only demonstrating a small set of tasks and objects. 

\subsubsection{Pre-trained models}
Open X-Embodiment addresses this issue by gathering a number of robotic datasets collected at many different labs, on different robots and in different environments, and assembling them into one unified dataset. While each individual dataset might be too narrow to train general purpose policies, the union of many such datasets can drastically improve coverage.
To evaluate the resulting dataset as well as the potential of positive transfer in robotics, the authors train several state-of-the-art models with the Open X-Embodiment data, published as pre-trained models.

\subsubsection{Dataset Composition}
At the time of publishing, the Open X-Embodiment dataset consisted of over 1M real robot trajectories, collected from 22 different robot embodiments. It was constructed by gathering 60 existing robot datasets from 34 robotic research institutions around the world. Since then, more datasets have already been added. All datasets are converted into a consistent format: The RLDS (Reinforcement Learning Datasets) format was used, which accommodates the various action spaces and input modalities of different robot setups. This format also allows for efficient data handling in all major deep learning frameworks.

\subsubsection{Pre-trained model RT-1-X}

One of the several state-of-the-art algorithms trained on this dataset is the RT-1 model, introduced in the previous section. This pre-trained model will further be used in this study. For a comparison of the performance of this model with other state-of-the-art models, see \cite{open-x-embodiment-collaboration-open-2024}.

\subsubsection{Action and Observation Space Alignment}
One challenge in training models on the heterogeneous X-Embodiment dataset is the significant variation in observation and action spaces across robots. To address this, a coarsely aligned action- and observation space is used across datasets: From each dataset, one camera view is selected, resized to a common resolution, and used as model input. Camera observations naturally vary substantially across datasets, as they are using different camera positions, scenes and embodiments. The dataset's action set is converted to a 7 DoF end-effector action and normalized, so the model's output can be interpreted differently depending on the embodiment. 

\subsection{UMI RTX Robotic Embodiment}
\label{sec:umi-rtx}

The robot used in this study is the RTX model from Universal Machine Intelligence Ltd. (UMI). 
It should be noted that after this section the UMI RTX robot will be referred to as the UMI robot or just the UMI, in order to avoid confusion with the RT-1-X model trained on the Open X-Embodiment.

\subsubsection{History and Use Cases}
The RT series is a family of robots that was first introduced in the mid 1980's, by the British company Universal Machine Intelligence Ltd., under the technical direction of Tim Jones. 
According to Jones, the UMI RTX was the first "co-bot" in existence\footnote{\href{https://www.linkedin.com/in/tim-jones-8464751/recent-activity/all/}{Tim Jones, Owner Intelligent Machines Ltd, LinkedIn post, 2021}}, as it was capable but still safe to use around humans due to its low-powered motors and belt drives, which were meant to act as mechanical fuses \cite{bassily-mechatronics-2007}. 
Right after its introduction, the UMI RTX was widely adapted in research settings, with multiple institutions using it for teaching  and experimentation with different technologies \cite{bauters_transputer-based_1992,  chintamani-comparing-2006, fewless-telerobotic-nodate, bassily-mechatronics-2007}. The UMI RTX was especially popular in the domain of rehabilitation and healthcare robotics.

After the 2000's, not much work was done with the UMI RTX robots. It is unclear how many of them are still in working condition. At KU Leuven, two master's theses were done in the years 2015 \cite{dooms_camera_2015} and 2016 \cite{nayer_camera_2016}, in which the authors worked with a UMI RTX robot that seems to be or have been in working order at KU Leuven.
Since 2022, there have been efforts to "revive" the UMI RTX at the  Intelligent Robotics Lab of the University of Amsterdam, which was originally used in 1992 to play chess \cite{groen_chess_1992}.
Control code that runs on modern Linux distributions 
was extended by \citeauthor{garde_ros_2023}, who built a ROS2 interface around the control code, as well as a graphical user interface to easily control the robot combined with object detection capabilities from depth images \cite{garde_ros_2023}.

\subsubsection{Technology}
The UMI RTX is a robot of the SCARA (Selective Compliance Assembly Robot Arm) type, which typically works on a fixed height but for in case of the UMI RTX is extended of vertical travel of the arm \cite{universal_machine_intelligence_ltd_inside_1987}. It has 7 degrees of freedom: The rotation of the elbow and shoulder joints, yaw, pitch, and roll of the wrist (end effector), the height of the entire arm assembly (z axis), and the opening of the gripper. Figure \ref{fig:umi-uva} shows the UMI robot at the Intelligent Robotics Lab, which will be used for this study. Due to its SCARA layout, the range of motion of the robot is kidney-shaped. A specific design choice is that the elbow joint and the wrist's yaw joint are driven through a combined spindle, meaning the wrist yaw always stays the same in relation to the robot base when moving the elbow joint. 


\section{Method}

Because initial experiments \cite{Salzer2024thesis} showed that zero-shot generalization was not possible to the unseen UMI robotic embodiment, the decision was made to fine-tune the model. 


\subsection{UMI dataset for Imitation Learning}
To evaluate the generalization potential of RT-1-X, it was fine-tuned on a single scenario from the UMI environment, meaning one task, one object, and one workspace setup. As with the zero-shot evaluation, object and task are chosen based on the ones that are already represented in Open X-Embodiment: The task is simply picking up an object from the workspace, the target object being a banana, which is amongst the most common items in Open X-Embodiment

For the workspace setup, values are chosen that led to large amounts of movement in the zero shot evaluation runs. As discussed in \cite{Salzer2024thesis}, the camera position was the variable which seemed to have the largest effect on the performance, so for the demonstration the "Side" position is chosen.

To demonstrate task execution on the UMI, a mode for manually controlling the robot needs to be established. For this experiement we have chosen for a PlayStation controller, which allows for intuitive, efficient control. A further discussion of the teleoperation strategy along with a comparison to other datasets from Open X-Embodiment can be found in an Appendix of \cite{Salzer2024thesis}.

\subsubsection{Data Processing}
On the software side, the aim was to collect demonstrations that are similar in format to the datasets in Open X-Embodiment, in order not to increase the embodiment gap further than necessary. 
The aim was therefore to create episodes of around 30 steps, which is the average of BridgeData, the biggest relevant dataset \cite{pmlr-v229-walke23a}. To achieve that with the UMI robot, which is very slow in movement compared to other embodiments, actions were logged every five seconds, or at a frequency of 0.2Hz. 


After each demonstration run, an episode is saved in a simple NumPy file format. RT-1-X uses Google's RLDS dataset format. One of the researchers behind Open X-Embodiment published a tool to build such datasets\footnote{\url{https://github.com/kpertsch/rlds_dataset_builder}}, this code was adapted and used to convert the demonstrations to the correct format.

To prevent any issues that stem from false or inaccurate action recording, code was written that loads the demonstration episodes one-by-one and replays the actions in a simulated environment. This allowed to inspect each demonstration run and to make sure that the correct movements were stored.

\subsection{Fine-Tuning RT-1-X}


There are several approaches to fine-tuning Transformer based models that only fine-tune certain layers, or even add additional layers to the pre-trained model that are then fine-tuned \cite{octo_model_team_octo_2024}. For this study, it was experimented with freezing the first Transformer layers and fine-tuning only the last 3 of the total 8 Transformer blocks. This did not seem to make a difference in outcome, it was decided to fine-tune the whole model for further experimentation.

\subsection{Hyperparameters}
There are multiple variables that have an influence on the training process, for which it is necessary to find suitable values. As a baseline, the values from the inital training code are used. By changing the values based on suggestions from literature, it is investigated if the fine-tuning process can be optimized.

\subsubsection{Batch size}
A batch size of 1024 was used during the Open X-Embodiment pre-training. Using a batch size of this size is not possible due to memory constraints: The machine available for training has a vRAM of 24GB. Additionally, it is common practice when fine-tuning foundation models to use a batch size much smaller than the one used during the initial training, as the fine-tuning dataset is much smaller \cite{howard-ruder-2018-universal}. By experimentation, it was found that the largest possible batch size on the available hardware is five episodes. This is because each episode contains multiple images, together with language embeddings and output actions.

\begin{figure}
    \centering
    \includegraphics[width=0.75\linewidth]{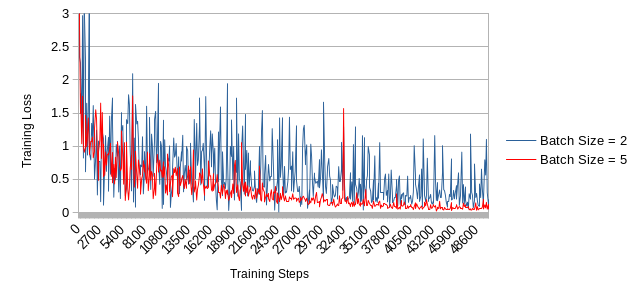}
    \caption{Using different batch sizes in training had significant effects on the noise in the loss curves. Compared here are training loss curves for batch size 5 (red) and batch size 2 (blue), with all other parameters unchanged. The biggest usable batch size in this research is five, due to hardware limitations.}
    \label{fig:batch-size}
    \vspace*{-1em}
\end{figure}

When comparing the training process with a batch size of 2 and a batch size of 5, it shows that the training is quicker and more constant with a bigger batch size. Figure \ref{fig:batch-size} shows a comparison of the training loss curves. It shows much less noise in the training curve of the higher batch size, suggesting that training could be further improved with a higher batch size.


\subsubsection{Learning rate}
\begin{figure}
    \centering
    \includegraphics[width=1\linewidth]{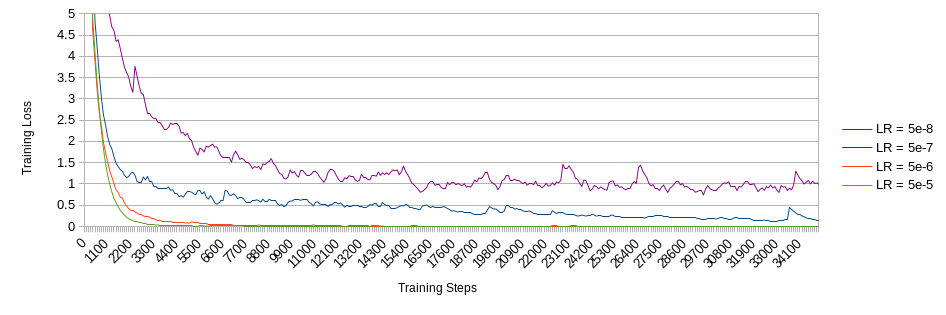}
    \caption{The choice of the learning rate proved to be essential for the quality of the training process. The chart shows the comparison of different learning rates when fine-tuning RT-1-X on the UMI dataset. A learning rate of 5e-6 was ultimately chosen.}
    \label{fig:lr-compare}
    \vspace*{-1em}
\end{figure}

One of the main parameters of the training process is the learning rate. RT-1 uses the Adam optimization algorithm \cite{kingma_adam_2017} for training. One of the main contributions of this algorithm is the dynamic adaptation of the learning rate, which leads to it typically requiring little tuning of the hyperparameters \cite{kingma_adam_2017}. In the published RT-1-X training code, the learning rate is set to 1e-4. Looking at other Transformer-based foundation models, the learning rate for fine-tuning is typically lower than the one used for initial training \cite{devlin_bert_2019}. It was experimented with a variety of different learning rates, the training loss curves of some of them are shown in Figure \ref{fig:lr-compare}. Ultimately, a learning rate of 5e-6 was chosen, which led to the most success.

\subsubsection{Training Steps}
Finally, it is important to train the model for a suitable amount of steps, allowing it to learn the behavior of the task without overfitting the training data. Through experimentation and comparison to similar fine-tuning setups, it was established that the best performance could be achieved at around 50.000 steps. Notably, the training process is quite noisy due to the encountered batch size limitations. To achieve the best possible performance, a training checkpoint was saved every 1000 training steps. Before running the experiments, all checkpoints from step 40.000 to step 60.000 were evaluated. The best checkpoint was found to be at 55.000 training steps, this was then used to run the experiments.

\subsection{Validation}
\begin{figure}
    \centering
    \includegraphics[width=1\linewidth]{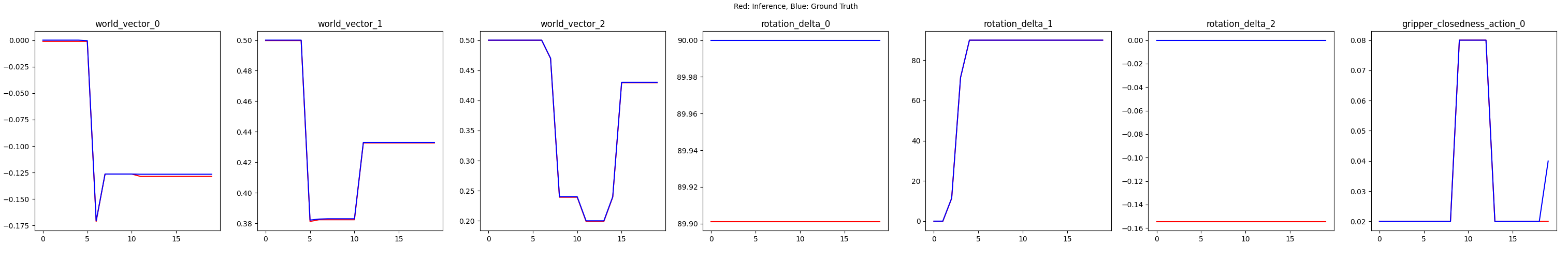}
    \caption{Inference is run with images from the UMI fine-tuning dataset to verify that fine-tuning was effective. Output of RT-1-X run with images from the UMI dataset is shown in red, compared to actions recorded during demonstration (ground truth) in blue.}
    \label{fig:validation-finetuned}
    \vspace*{-1em}
\end{figure}

To verify that the fine-tuning process was successful, inference is run with the images and language instruction from an episode of the training dataset. Figure \ref{fig:validation-finetuned} shows the original actions in the recorded episode (blue) and the predicted actions by the fine-tuned model (red). 

\section{Results}
\label{sec:results}

The first step of this study is to investigate if the performance could be improved by fine-tuning the model on demonstrations from the UMI robot. The RT-1-X model was fine-tuned with a dataset of 100 manual demonstrations of the UMI executing a simple task (picking up a banana from the workspace).

\subsection{In-Distribution Performance}
The first experiment evaluated the performance of the fine-tuned model on the demonstrated task. In this experiment, a success rate of 23\% could be achieved. In most of the non-successful evaluation runs, the robot executed the correct movements, and was off from the target object by less than five centimeters. Experiment runs where the robot executes the correct sequence of actions and is off by less than five centimeters are classified as "near miss". All evaluation runs that resulted either in success or a near miss combined make up 80\% of all evaluation runs. Table \ref{tab:res_finetuned} shows the outcome of the experiments.

\begin{table}[h!]
    \centering
    \begin{tabular}{|c|c|} \hline 
         \textbf{Observation}& \textbf{Amount}\\ \hline
         Success& 7\\ \hline 
         Near Miss& 17\\ \hline 
         Failure& 6\\ \hline 
         Total& 30\\ \hline
    \end{tabular}
    \caption{Results of the fine-tuned model on the demonstration task}
    \label{tab:res_finetuned}
    \vspace*{-1em}
\end{table}

Notably, the model shows signs of being biased towards certain positions on the workspace. When repeating a near miss run with the target object position unchanged, the robot typically executes a near miss in the exact same position. When moving the target object to this position, the robot is able to successfully pick it up. Possible reasons for this behavior are discussed in Section \ref{sec:discussion}.

\subsection{Transfer of Object Knowledge}

The second experiment aimed to investigate if objects that were only seen in Open X-Embodiment during pre-training are recognized on the UMI, by the fine-tuned model. A coke can, which is a common object in Open X-Embodiment, was placed on the workspace and the model was instructed to pick it up. As a second part to this experiment, the coke can and the original banana were placed on the workspace, the robot had to decide which one to pick up.

\begin{table}[h!]
    \parbox{.45\linewidth}{
    \centering
    \begin{tabular}{|c|c|} \hline  
          \textbf{Observation}& \textbf{Amount}\\ \hline 
            Success&1\\ \hline 
          Near miss& 5\\ \hline  
          No attempt& 4\\ \hline  
        Total&10\\ \hline 
    \end{tabular}
    \caption{Results of can pick up with only can in workspace}
    \label{tab:res_can_only}
    }
    \hfill
    \parbox{.45\linewidth}{
    \centering
    \begin{tabular}{|c|c|} \hline 
          \textbf{Observation}& \textbf{Amount}\\ \hline 
          Coke pick up attempt& 5\\ \hline 
          Banana pick up attempt& 3\\ \hline 
          No attempt& 2\\ \hline 
 Total&10\\ \hline
    \end{tabular}
    \caption{Results of can pick up with can and banana in workspace}
    \label{tab:res_can_banana}
    }
    \vspace*{-1em}
\end{table}

Table \ref{tab:res_can_only} shows the results of the experiment with only the can present in the workspace. A success rate of 10\% and a success or near miss rate of 60\% was achieved. When both the coke can and the banana object were placed on the workspace, the results were significantly worse, with no successful pick ups. To capture the object recognition abilities of the model, these experiments are classified as either near miss attempts on the coke can, near miss attempts on the banana, or no significant attempts at either of the objects. The results of this are shown in Table \ref{tab:res_can_banana}.

\section{Discussion}
\label{sec:discussion}

When evaluated on the in-distribution task, the fine-tuned model shows a success rate of 23\% over 30 experiment runs. This is a significant achievement compared to the non-existent zero-shot performance and shows that it is possible to further improve the RT-1-X model for specific embodiments by fine-tuning. 

This number is however still low when compared to the performance of RT-1-X on embodiments that were part of its pre-training. In the experiments reported on the Open-X embodiment there  settings with comparable success rates (e.g. Berkeley Bridge with a success rate of 27\% \cite{pmlr-v229-walke23a}), but most of the experiments reported higher success rates (e.g. Freiburg Franka Play with a success rate of 72\% \cite{mees_what_2022}).

When looking at the failed evaluation runs, the test results show that 77\% of all failed attempts are near misses, meaning the robot executed the correct sequence of actions to pick up the target object, however it was off on the X and Y coordinates by a small margin. This means that 80\% of evaluation runs led to a result of either success or near miss, showing that while the model managed to pick up the behavior patterns of the UMI robot, it is struggling with the accuracy of its movements. 



The second experiment investigated if the model recognizes objects that were only seen in pre-training. After initial success with using a coke can as the target object instead of the banana, an additional experiment was conducted to investigate the models object recognition capabilities: When running the pick up task with both the banana and the coke can on the workspace, the model did not show capabilities of reliably identifying the correct object. In around half the attempts, the robot tried to pick up one of the two object, the other half of the experiments did not show any significant pick up attempts. This implies two things: First, knowledge about other objects is not transferred from the pre-training data. The second, unexpected observation is that the model is also not able to reliably identify the object it has been fine-tuned on as soon as there is another object on the table. The model has only learned to pick up an object from the workspace, but has no further understanding of what the object is.

Note that RT-1 model of Google is in July 2023  followed up with the RT-2 model \cite{brohan_rt-2_2023} (which is unfortunately not open-source). In May 2024 Octo \cite{octo_model_team_octo_2024} was introduced by many of the researchers behind Open X-Embodiment, a model optimized to be fine-tuned. In June 2024 OpenVLA \cite{kim2024openvlaopensourcevisionlanguageactionmodel} was introduced, which again outperforms RT-1-X, Octo and RT-2-X. Yet, all those initiatives are based on the robots with a spherical workspace in the Open X-Embodiment dataset, so generalization to a SCARA robot like the UMI remains an open issue.

\quad \linebreak
\quad \linebreak

\section{Conclusion}
\label{sec:conclusion}

This study explored the generalization capabilities of the RT-1-X model, particularly its ability to adapt to a robot type not seen before.  A dataset of demonstrations was collected using the UMI robot, and a fine-tuning pipeline for RT-1-X was developed. The fine-tuning process effectively enhanced the model's performance on the new embodiment and the learned task, although the performance did not reach the levels achieved on the embodiments seen during pre-training. Additionally, it was found that no concrete knowledge about objects to be manipulated was transferred from the pre-training dataset.

\printbibliography[heading=bibintoc]

\end{document}